\documentclass[runningheads]{llncs}
\usepackage{graphicx}
\usepackage{xcolor}
\usepackage{subcaption}
\usepackage{booktabs}
\usepackage{cite}
\usepackage{amsmath,amssymb,amsfonts}
\usepackage{algorithmic}
\usepackage{textcomp}
\usepackage{threeparttable}
\begin{document}
\title{Hub and Spoke Logistics Network Design for Urban Region with Clustering-Based Approach
}
\titlerunning{Logistics Network Design for Urban Area}
%

\author{
    Quan Duong\orcidID{0000-0002-4493-2331} \and
    Dang Nguyen\orcidID{0000-0002-3664-4364} \and
    Quoc Nguyen\orcidID{0000-0002-1692-1918}
}
%
\authorrunning{Quan Duong et al.}
%
\institute{
GHN Data Science, Ho Chi Minh, Vietnam \\
\email{\{quandb,dangnh1,quocnd\}@ghn.vn}}

\maketitle              
\begin{abstract}
This study aims to propose effective modeling and approach for designing a logistics network in the urban area in order to offer an efficient flow distribution network as a competitive strategy in the logistics industry where demand is sensitive to both price and time \cite{cheong2005logistics}. A multi-stage approach is introduced to select the number of hubs and allocate spokes to the hubs for flow distribution and hubs' location detection. Specifically, a fuzzy clustering model with the objective function is to minimize the approximate transportation cost is employed, in the next phase is to focus on balancing the demand capacity among the hubs with the help of domain experts, afterward, the facility location vehicle routing problems within the network is introduced. To demonstrate the approach's advantages, an experiment was performed on the designed network and its actual transportation cost for the real operational data in which specific to the Ho Chi Minh city infrastructure conditions. Additionally, we show the flexibility of the designed network in the flow distribution and its computational experiments to develop the managerial insights which contribute to the network design decision-making process.

\let\thefootnote\relax\footnote{Proceedings of the 34th International Conference on Industrial, Engineering \& Other Applications of Applied Intelligent Systems (IEA/AIE 2021), Kuala Lumpur, Malaysia. Copyright 2021 by the author(s)\hfill}

\keywords{Logistics network design  \and Hub-and-spoke \and Clustering.}
\end{abstract}
\section{Introduction}\label{intro}
The volume of express delivery gets increasing with the rapid development of e-commerce. At the same time, intense competition has forced express delivery service companies to compete in different categories to get ahead in the race. Transportation cost and delivery lead-time are considered to be the most important metrics for success. On the other hand, logistics network design is concerned with the number of locations of stations that allocate for customer's demand points, and the assignment of the stations to the distribution centers within their coverage. The optimal setting must distribute the goods to the customers with the least cost and satisfy the service level agreement. To illustrate these strategies, considering the current configuration of the delivery network of logistics providers in Vietnam, an attempt to design a network that distributes more efficiently and reduces the transportation cost. Thus, the primary goal of this work is to, firstly, examine the limitation of the single distribution center. Secondly, design a delivery network that optimizes the transportation cost by demonstrating for an urban region. Moreover, we develop an analytical framework that offers multiple distribution scenarios within the designed network and its impact on the operational performance of the firm. 

The rest of the paper is organized as follows: In Section \ref{lr}, the relevant studies on logistics network design for the urban region are reviewed. Afterward, Section \ref{pa} presents our practical approach and framework and establishes the flow distribution scenarios within the network. The experiment results take into account the real operational data from a delivery network with a case study for Ho Chi Minh city will be conducted in Section \ref{er}. Finally, we conclude the main findings and future research directions in Section \ref{cnf}.

\section{Literature Review}\label{lr}
The logistics network design problem consisting of multiple facilities location problem \cite{KUCUKDENIZ20124306}, each facility has a limit of serving demand points contribute the difficulty of problem where need to solve the single facility location problems, but also searching for the optimal number of facilities and assigning the customer demand points for each cluster. Additionally, Cheong et al \cite{cheong2005logistics} explored the benefit of segmenting demand points into different classes, they assumed that each demand point can be divided into two classes of the sensitivity to delivery lead time, the segmentation results in more effective allocation in demand point and reduce the network cost. A typical use of flow distribution, utilizes the distribution centers (DC) to serve the customer's demand points within their coverage. In a single DC scenario, vehicles depart from DC's location and deliver to the customer's demand points, this process makes the logistics resource difficult to be shared in high-density areas. Moreover, the logistics service provider manages to accommodate customer's demanding at time window and price make the operations more challenging. In 2016, Roca-Riu et al \cite{roca2016evaluation} took a further step to evaluate the urban good distribution for several European cities with a single independent DC and proposed continuous models to improve the efficiency of urban distribution with the use of urban consolidation centers (UCC), the quantitative metric indicated that by deploying the UCC can reduce the transportation cost up to 14\%. The collaborative and multi-depots in urban areas got a lot of attention in recent years \cite{LONG201684,WANG2017143,XU2017684}. Later in 2020, Wang et al \cite{WANG2020112910} solve collaborative multi-depots vehicle routing problems to deal with the impact of changing time window. In order to group the demand points into a single, weighted K-mean clustering and Fuzzy C-means clustering algorithms stand out for the facility location problem \cite{esnaf2009fuzzy,KUCUKDENIZ20124306,hu2020network}. However, Meng et al \cite{you2019optimal} figured out that weighted K-mean always traps into local optima and is sensitive to the initial setting, and very limited research incorporate the domain knowledge for the target region with fuzzy c-means clustering capacity.

In this study, the fuzzy c-mean (FCM) clustering algorithm \cite{bezdek1984fcm} is employed to solve the uncapacitated multi-facility location problem. The FCM allows the data point belongs to multiple clusters with different degrees of membership. Combining the degree of membership features and domain expert for the target region of designing the delivery network, the belonging of demand point to the cluster will be judged by humans in order to balance the capacity among the clustering and other area's characteristics.

\section{Methodology}\label{pa}

Given all the spokes (know as the demand points that will serve the end customers for a specific area). The methodology aims to find the optimal  Hub's Location and spoke to the hub assignment with respect to minimize the transportation cost. The process consisting of three phases: i) Clustering: detecting the group of nearby spokes to a cluster. ii) Hub's location: detect the cluster's center in which serve as the local DC. iii) Vehicle Routing: determine the flow distribution scenario among the Hubs' network, and perform the vehicle routing for good delivery to the spokes. 
\subsection{Clustering}
\label{sub-sec:clustering}
Formally, a dataset \textit{X} as a set of \textit{N} demand points represented as vectors in an \textit{2}-dimensional space: $X = \{x_1, x_2, ..., x_N\} \subseteq \Re^{\textit{2}}$. A clustering in \textit{X} is a set of disjoint clusters that partitions \textit{X} into \textit{K} groups: $C = \{c_1, c_2, ..., c_k\}$ where $\cup_{c_k \in C} c_k = X, c_k \cap c_l = \emptyset  \forall k \ne l$. The Haversine distance between objects $x_i$ and $x_j$ will denote as $d_h (x_i, x_j)$. The approximate transportation cost is calculated as
\begin{equation}\label{cost}
    Cost = \sum_{c_k \in C} \left\{ \sum_{c_l \in C \setminus c_k} d_h(\overline{c_k}, \overline{c_l}) + \sum_{x_i \in c_k} d_h(\overline{c_k}, x_i) \right\}
\end{equation}

\begin{equation}\label{centriod}
    \overline{c_k} = \frac{\sum_{x_i \in X} {w_k(x_i)}^m * x_i}{\sum_{x_i \in X} {w_k(x_i)}^m}
\end{equation}
\noindent Where $\overline{c_k}$ is the centroid of a cluster $c_k$ is defined by the FCM, \textit{m} is the hyper-parameter that controls the fuzziness of clustering and $w_k(x)$ is the degree of membership that data point \textit{x} belong to the cluster \textit{k}.

\subsubsection{Final Cluster Assignment}
The degree of membership feature from FCM is utilized for balancing the demand among the cluster together with humans. Specifically, taking into account the domain knowledge of the target designed area, those mixture-membership demand data points will be re-assigned with the domain expert in the loop. Based on the FCM's degree of membership feature, the expert will assign each single demand data point into a target cluster in order to satisfy the capacity balancing requirements. The number of clusters and algorithm parameters is searching by incremental methods with the objective is to minimize the approximate transportation cost in the formula \ref{cost}. In the Section \ref{sub-section:detected_clusters}, the experiment results with specific parameters will be conducted.

\subsection{Hub's location}
\label{sub-section:hub_location}
After allocating of demand data point to cluster, the center of gravity method \cite{ballou1999business} is employed to determine the optimal Hub's location. The aims of center of gravity method is to bring the center of cluster closer to demand points with higher demand. For all $\textit{k} \in \{1,2, ..., K\}$ and $\nu(x)$ is the delivery demand for demand point \textit{x}, the center gravity of each cluster \textit{k} are calculated as:
\begin{equation}\label{centriod}
    {h_k} = \sum_{x \in c_k} {\beta(x) * x}
\end{equation}
\begin{equation}\label{centriod}
    \beta(x) = \frac{\nu(x)}{\sum_{x \in c_k} {\nu(x)}} 
\end{equation}

The $\beta(x)$ value serves as the weighted factor within the cluster. The larger value will pull the hub's location to its side. The real transportation cost corresponding to these locations will be shown in Table \ref{table:actual_cost}.
\subsection{Flow Distribution and Vehicle Routing}
\label{sub-section:flow_dist_n_veh_routing}
\begin{figure}[htbp!]
\includegraphics[width=\textwidth]{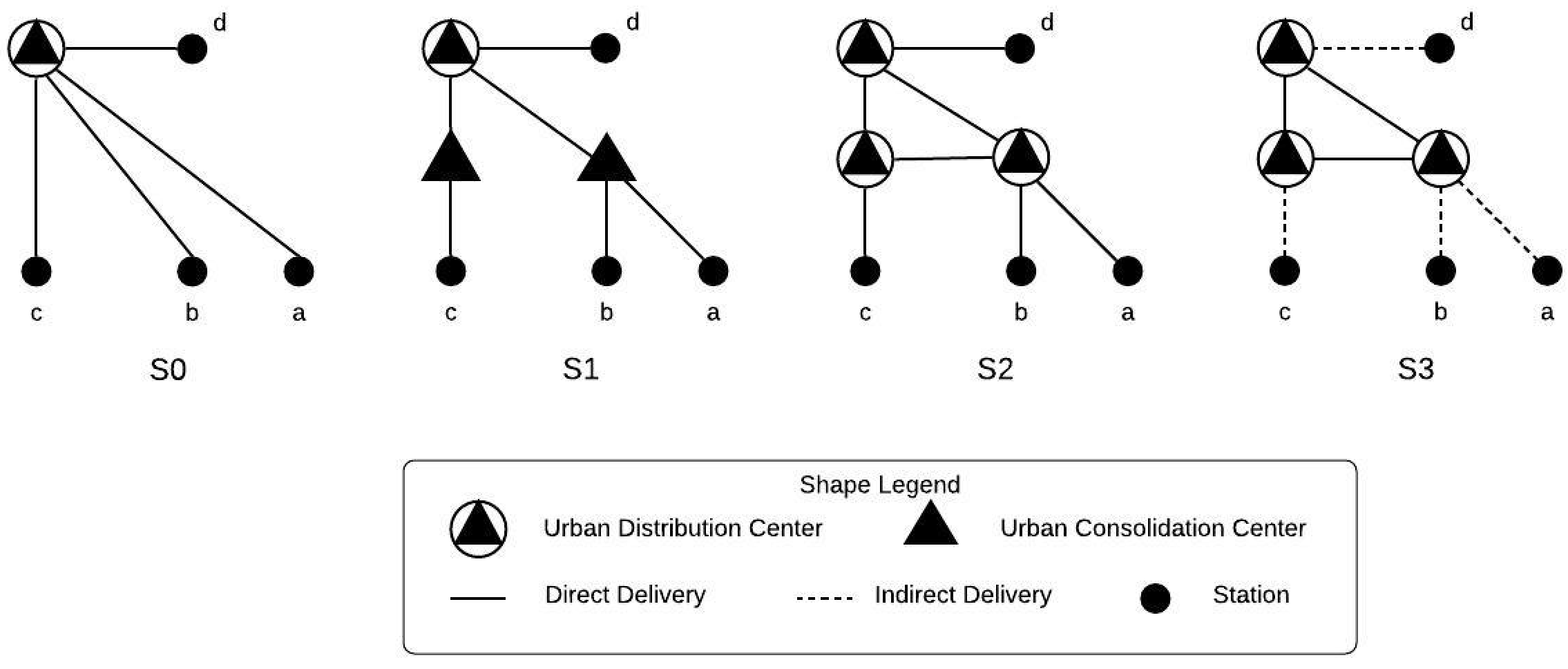}
\caption{Flow Distribution Scenarios}
\label{fig:flows}
\end{figure}
Given a designed network, the authors deliberately prototype three flow distribution scenarios within the network in the Figure \ref{fig:flows}, and a default scenario which is using for daily operation in GHN network as follows: $S0$: Default scenario, single independent DC. $S1$: Each Hub will serve as Urban Consolidation Center (UCC). $S2$: Each Hub will serve as Urban Distribution Center (UDC). $S3$: Each Hub will serve as UDC, and drop the delivery packages at its center for the nearby demand points. Here, the newly designed network has the ability to offer three different flow distribution scenarios. Fundamentally, $S1$ only performs the loading/unloading packages to the designated DC. An upgrade version of $S1$, $S2$ also has the ability to sort the packages within its coverage and distribute all other packages to other DC/UDC. As an advantage on multi-depot, the DC locate closer to its demand points, $S3$ only need to sort packages, and hand-off the packages at its location, this scenario offers flexibility for the urban area such as reduce the cost for delivering by trucks, restricted during the rush hours. 

\section{Experiment Result}\label{er}
The numerical results reported in this paper are the real operational data in Ho Chi Minh City, Vietnam. Including, i) \textit{77} active demand data points. ii) All transactions data for the entire December 2020 has been extracted and calculated for two different phases in this paper. First, the average delivery demand for the whole period would be used for representing the demand point's volume that contributes to the Hub's Location inference in Section \ref{sub-section:hub_location}. Second, transaction data of \textit{$22^{nd} December$} has been used to conduct experimental design for pickup and delivery demand in the designed network. Essentially, the actual transportation cost in Section \ref{sub-section:actual_trans_cost} are computed based on this day.


\subsection{Detected Clusters}
\label{sub-section:detected_clusters}
Regarding to experimental setting, the authors employed the Python's scikit fuzzy package\footnote{https://pythonhosted.org/scikit-fuzzy/} and the configuration ($\textit{m} = 3$, $\textit{error} = 0.002$, $\textit{maxiter} = 1000$, $\textit{seed} = 12345$, and $\textit{c} = \{2,3,4,5\}$). The primary parameter is \textit{c} (the desired number of clusters). The best parameter in determined by the combination of approximate transportation cost in Formula \ref{cost}, fuzzy partitioning coefficient, and the variation of delivery demand for each cluster. 

\begin{table}[ht!]
\caption{Summary of Clustering Results}
\centering
\label{table:cluster_sum}
\begin{tabular}{l l l}
\toprule
\multicolumn{1}{l@{\hskip 0.2in}}{Number of clusters} & \multicolumn{1}{l@{\hskip 0.2in}}{Approximate transportation cost} & \multicolumn{1}{l}{Coef}\\
\midrule
2 & 527.8 & 0.568\\
3 & \textbf{482.8} & 0.424\\
4 & 490.3 & 0.362\\
5 & 530.6 & 0.308\\
\bottomrule
\end{tabular}
\end{table}

\begin{figure}[ht!]
\includegraphics[width=.9\textwidth]{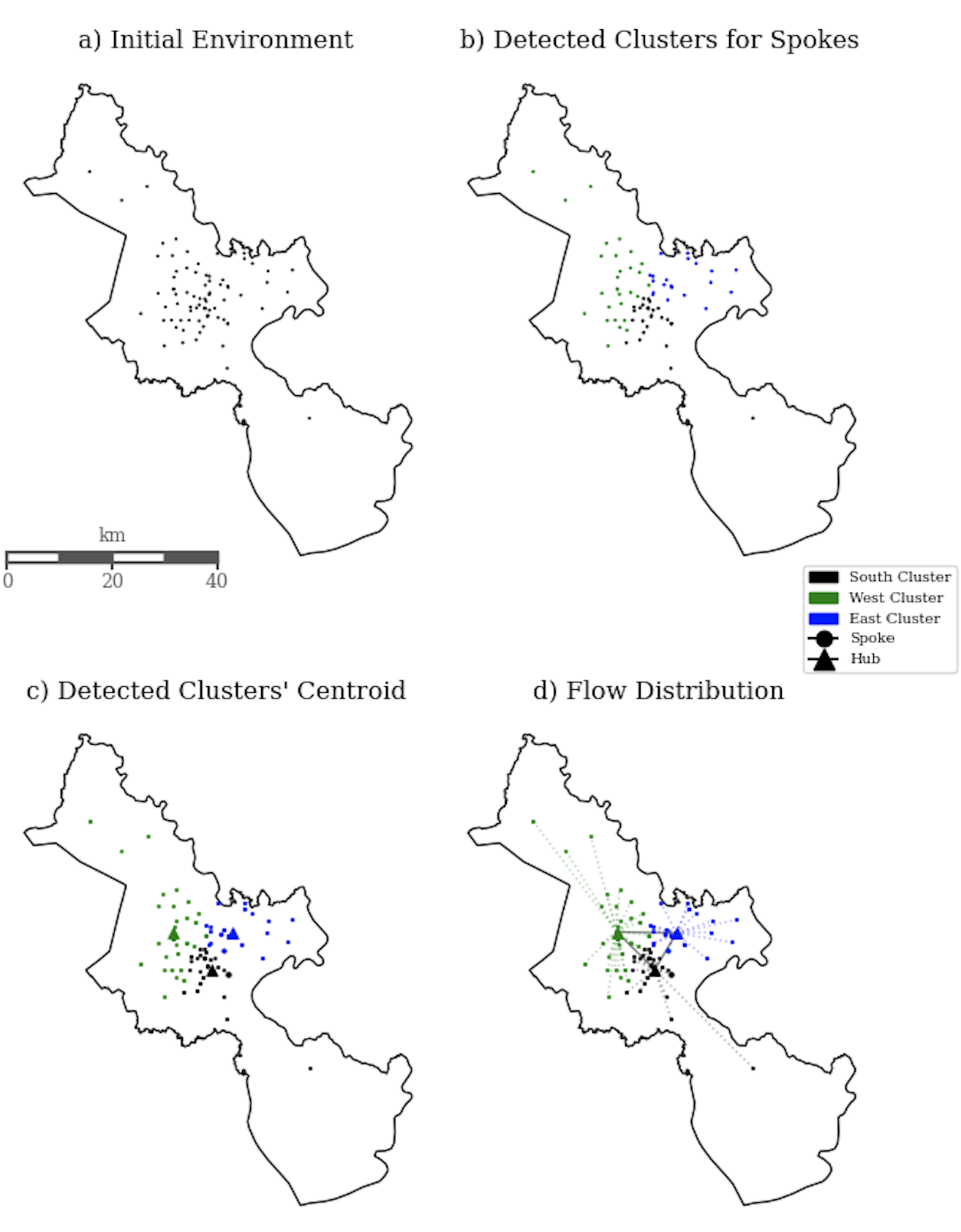}
\caption{The environment and its clusters between the elements in network.}
\label{detected_clusters}
\end{figure}

The experimental results of the clustering step shown in Table \ref{table:cluster_sum}, The transportation cost with \textit{3} and \textit{4} show the better result than the other values. Take a deeper look, we can see that the \textit{3} clusters's cost and \textit{Coef} slightly better than the \textit{4}. Inspiring by these numbers, the exact assignment of data points to the cluster will be adjusted by the domain expert due to the fact that it's mandatory the satisfy the capacity requirements. Figure \ref{detected_clusters} shows the visual outcome of each step for the entire process. The initial setting of \textit{77} demand points display as black dots belong to the Ho Chi Minh region in Sub-figure \ref{detected_clusters}-a, each color in the right next Sub-figure \ref{detected_clusters}-b represents the 3 detected clusters by virtue of the FCM algorithm and human judgment. A single triangular in Sub-figure \ref{detected_clusters}-c indicates the Hub's location for each cluster. The right-bottom Sub-figure \ref{detected_clusters}-d visually shows the flow distribution from Hub to Hub and from Hub to Spokes within sub-network.

\subsection{Actual Transportation Cost}
\label{sub-section:actual_trans_cost}
Confidentially, with the goal of comparing the transportation cost among the scenarios, also not to disclose the commercial confidentiality information. The values for \textit{number of trucks} and \textit{total cost} for the current operation mode have been changed to \textit{v} and \textit{c} respectively. The actual transportation cost specified in the Formula \ref{cost}. Due to the fact that the actual situation takes into account many important and realistic factors like delivering demand at each demand point, truck capacity, time window requirement, local traffic condition. Thus, we leverage the capacitated vehicle routing with time window solutions. In this regard, the optimization software OR-Tool \cite{ortools} is used to perform the computational step given the operational constraints and the conditions of Ho Chi Minh city such as travel distance and duration between points, time window requirement, delivery demand, and truck's capacity.

\begin{table}
\caption{Actual Transportation Cost for Scenarios}
\centering
\label{table:actual_cost}
\begin{threeparttable}
\begin{tabular}{l@{\hskip 0.5in} l@{\hskip 1in} l@{\hskip 0.5in} l@{\hskip 0.5in} l@{\hskip 0.5in}}
\toprule
\multicolumn{1}{l}{Scenario} & \multicolumn{1}{l}{Number of trucks} & \multicolumn{1}{l}{Total cost} & \multicolumn{1}{l}{Pickup cost} & \multicolumn{1}{l}{Delivery cost}\\
\midrule
$S0$ & $v$ \tnote{*} & $c$ \tnote{**} & 0.7c & 0.3c\\
$S1$ & 0.79$v$ & 0.75$c$ & 0.44$c$ & 0.31$c$\\
$S2$ & 0.8$v$ & 0.72$c$ & 0.47$c$ & 0.25$c$\\
$S3$ & 0.63$v$ & 0.61$c$ & 0.46$c$ & 0.15$c$\\
\bottomrule
\end{tabular}
\begin{tablenotes}\footnotesize
\item[*] Number of trucks need for the \textit{S0} scenario
\item[**] The total amount of money take for the \textit{S0} scenario
\end{tablenotes}
\end{threeparttable}
\end{table}

In the Table \ref{table:actual_cost}, we shown the actual transportation cost for all the mentioned scenarios. By introducing the multi-depot in the delivery network, the transportation cost reduces 25\%, 28\% by adding consolidation and distribution centers respectively. Take a deeper look in the \textit{S2} scenario, we can observe that the delivery cost is cheaper than \textit{S1} scenario 19\% by reducing the traveled distance for the delivery trips, \textit{S2} also takes more effort in the pickup trips in order to move the packages to closer DC. On the other side, about 20\% of the number of trucks have been reduced for performing delivery with the same demand in the network.

\section{Conclusions}\label{cnf}
This paper analyzes the advantages of collaborative multi-depot settings in the urban area, applied the iterative approaches for the logistics network design problem. By employing the FCM algorithm for clustering and leveraging the degree of membership feature to reassign the demand point to the cluster in order to balance the facility's capacity in each specific sub-network by virtue of domain expert. Based on the real operational data in Ho Chi Minh city, the results indicate that 3 clusters achieve the best solution in approximate transportation cost and reduce the actual transportation cost to 28\% and 20\% trucks used with 3 distribution centers compared to the original independent DC. Increasing the truckload utilization for the delivery trips between Hubs not only reduces the transportation cost and improves operational efficiency for the firm, but also put an effort into alleviating the traffic congestion, reducing stress for transportation infrastructure. In the future, the studies may focus on two directions. First, calculate the system cost for the designed network including the cost for opening a new facility and operating that facility with a given demand, and perform the experiment for multiple urban regions in Vietnam. Second, schedule the dynamic vehicle routing to deal with the highest demand days and measure its performance. Prospective studies on these directions would contribute to operating more efficiently and resilient on both typical and high demand days, also the promotion of multi-depots setting for the urban regions.
%
%
\bibliographystyle{splncs04}
\bibliography{mybibliography}

\end{document}